\documentclass[11pt,a4paper]{article}
\usepackage[hyperref]{emnlp2020}
\usepackage{times}
\usepackage{latexsym}

\usepackage{graphicx}
\usepackage{enumitem}
\usepackage{float}

\usepackage{changepage}

\usepackage{todonotes}

\usepackage{microtype}

\aclfinalcopy 
\def\aclpaperid{3237} 

\title{The birth of Romanian BERT}

\author{
  Stefan Daniel Dumitrescu \\
  Independent Researcher \\
    Bucharest, Romania \\
  \\\And
  Andrei-Marius Avram \\
  Politehnica Univ. of Bucharest \\
  Bucharest, Romania \\
  \texttt{\{dumitrescu.stefan,avram.andreimarius,sampo.pyysalo\}@gmail.com}
  \\\And
  Sampo Pyysalo \\
  University of Turku \\
  Turku, Finland \\
}

\date{}

\begin{document}
\maketitle
\begin{abstract}
Large-scale pretrained language models have become ubiquitous in Natural Language Processing. However, most of these models are available either in high-resource languages, in particular English, or as multilingual models that compromise performance on individual languages for coverage.
This paper introduces Romanian BERT, the first purely Romanian transformer-based language model, pretrained on a large text corpus. We discuss corpus composition and cleaning, the model training process, as well as an extensive evaluation of the model on various Romanian datasets. We open source not only the model itself, but also a repository that contains information on how to obtain the corpus, fine-tune and use this model in production (with practical examples), and how to fully replicate the evaluation process.
\end{abstract}

\section{Introduction}

A revolution started in natural language processing (NLP) a few years ago, when the first Transformer-based model \cite{vaswani2017attention} demonstrated a significant increase in state-of-the-art results compared to previous neural approaches. The bidirectional BERT \cite{devlin2018bert} language model has since been widely adopted as the baseline for transformer models, and it has been successfully applied to a broad range of NLP tasks from standard language modeling to question answering, text summarization, and machine translation.

A number of papers have been dedicated to studying why and how this model performs so well, including comparison to classical NLP \cite{tenney2019bert}, investigation of the newly introduced multi-head attention mechanism \cite{michel2019sixteen}, and analyses of what BERT learns \cite{clark2019does}. A good recent review is presented by \newcite{rogers2020primer}.
Following BERT, a number of variations of language models using the transformer architecture have been introduced, including extended models such as XLM \cite{lample2019crosslingual} and XLNet \cite{yang2019xlnet} as well as more efficient ones like DistillBERT \cite{sanh2019distilbert}, ALBERT \cite{lan2019albert}, and ELECTRA \cite{clark2020electra}. 

\begin{table}
\centering
\setlength{\tabcolsep}{4pt}
\begin{tabular}{lrrrr}
\hline \textbf{Corpus} & \textbf{Lines} & \textbf{Words}  & \textbf{Size} \\ \hline
OPUS      & 55.1 M    & 635.0 M    & 3.8 GB      \\
OSCAR     & 33.6 M    & 1725.8 M  & 11 GB       \\
Wikipedia & 1.5 M     & 60.5M    & 0.4 GB      \\
\textbf{Total}     & \textbf{90.2 M}    & \textbf{2421.3 M}   & \textbf{15.2 GB}  \\
\hline   
\end{tabular}
\caption{\label{font-table} Corpus statistics (post-cleaning).}
\end{table}

However, the vast majority of these studies have focused only on English models. While Google has released a multilingual BERT model trained on 100+ languages, only recently have monolingual models for other languages started to appear: FlauBERT for French \cite{le2019flaubert}, BERTje for Dutch \cite{vries2019bertje} and FinBERT for Finnish \cite{virtanen2019multilingual}. But none for Romanian, until now. 
While the multilingual BERT can be used also for Romanian, a monolingual model can bring an increase in accuracy that can be observed also in downstream task performance. Thus, we here introduce {\small\verb|Romanian BERT|}. 
This paper focuses on the technical and practical aspects of {\small\verb|Romanian BERT|}, covering corpus composition and cleaning in Section~\ref{Corpus}, the training process in Section~\ref{Training}, and evaluation on Romanian data sets in Section~\ref{Evaluation}.

\section{Corpus}
\label{Corpus}

\begin{table*}[ht]
    \centering
    \begin{tabular}{ll}
            \hline
            \textbf{Model} & \textbf{Tokenized Sentence} \\
            \hline
                {\tiny\verb|M-BERT (uncased)|} & \small {cinci bici \#\#cl \#\#isti au pl \#\#eca \#\#t din cr \#\#ai \#\#ova spre so \#\#par \#\#lita .}  \\
                {\tiny\verb|M-BERT (cased)|} & \small {Ci \#\#nci bi \#\#ci \#\#cl \#\#i \#\#ști au pl \#\#eca \#\#t din C \#\#rai \#\#ova spre Ș \#\#op \#\#âr \#\#li \#\#ța .}  \\
                {\tiny\verb|Romanian BERT (uncased)|} & \small {cinci bicicliști au plecat din craiova spre șopâr \#\#lița .}  \\
                {\tiny\verb|Romanian BERT (cased)|} & \small {Cinci bicicliști au plecat din Craiova spre Șo \#\#p \#\#âr \#\#lița .} \\
            \hline
    \end{tabular}
    \caption{Model tokenization examples. Note that M-BERT uncased strips accents.}
    \label{tab:vocab-example}
\end{table*}

The unannotated texts used to pre-train {\small\verb|Romanian BERT|} are drawn from three publicly-available corpora: OPUS, OSCAR and Wikipedia.

\textbf{OPUS}~~ OPUS \cite{Tiedemann2012ParallelDT} is a collection of translated texts from the web. It is an open-source parallel corpus that was gathered automatically, without human curation. It contains diverse domains of text, from medical prescriptions to law articles and movie subtitles. In total, the OPUS corpus contains around 4GB of Romanian text.

\textbf{OSCAR}~~ OSCAR \cite{ortizsuarez:hal-02148693}, or Open Super-large Crawled ALMAnaCH coRpus is a huge multilingual corpus obtained by language classification and filtering of the Common Crawl corpus. The Romanian section has about 11GB of text. It contains de-duplicated shuffled sentences.

\textbf{Wikipedia}~~ The Romanian Wikipedia is publicly available for download. We used the February 2020 Wikipedia dump, which contained approx.\ 0.4GB of text after cleaning.

\vspace{0.25cm}
\noindent
All corpora were subjected to the same cleaning procedure, with Wikipedia also needing extra cleaning as the XML extraction still contained markup tokens. The cleaning involved several sequential steps: 
\begin{itemize}[noitemsep,topsep=0pt,partopsep=0pt,parsep=0pt]
\item Remove all lines under a minimum length.
\item Remove all non-UTF8 tokens and all lines that contain forbidden characters / too many numbers (over 25\%) / non-ASCII characters (over 40\%) / too few letters (under 50\%).
\item Correct badly hyphened words (e.g. "a- mi" to "a-mi"), badly hyphened measure units (e.g. "km/ h" to "km/h"), badly formatted numbers (e.g. "12, 5\%" to "12,5\%").
\item Remove soft hyphens, URLs, emails.
\item Normalize dashes (there are several types of Unicode dashes) and other characters.
\item Reduce multiple spaces.
\end{itemize}

\section{Pretraining} \label{Training}

The pretraining process starts with building a vocabulary on the available corpus. Using  byte-pair-encoding (BPE), we generated cased and uncased vocabularies containing 50000 word pieces. Character coverage\footnote{To reduce UNKs we use a larger character set to cover less frequent chars/symbols} was set to 2000.

\begin{table}[ht]
    \centering
    \begin{tabular}{lcc}
          & \textbf{Tokens/} & \textbf{UNK/} \\
          \textbf{Vocabulary} & \textbf{Word} & \textbf{Word}  \\
          \hline
            {\small\verb|M-BERT (uncased)|} & 2.00 & 0.005 \\
            {\small\verb|M-BERT (cased)|} & 1.82 & 0.004  \\
            {\small\verb|Romanian BERT (uncased)|} & 1.41 & 0.0003  \\
            {\small\verb|Romanian BERT (cased)|} & 1.37 & 0.0002 \\
          \hline
    \end{tabular}
    \caption{Vocabulary statistics.}
    \label{tab:vocab-stats}
\end{table}

\begin{figure}[!ht]
    \centering
    \includegraphics[width=0.45\textwidth]{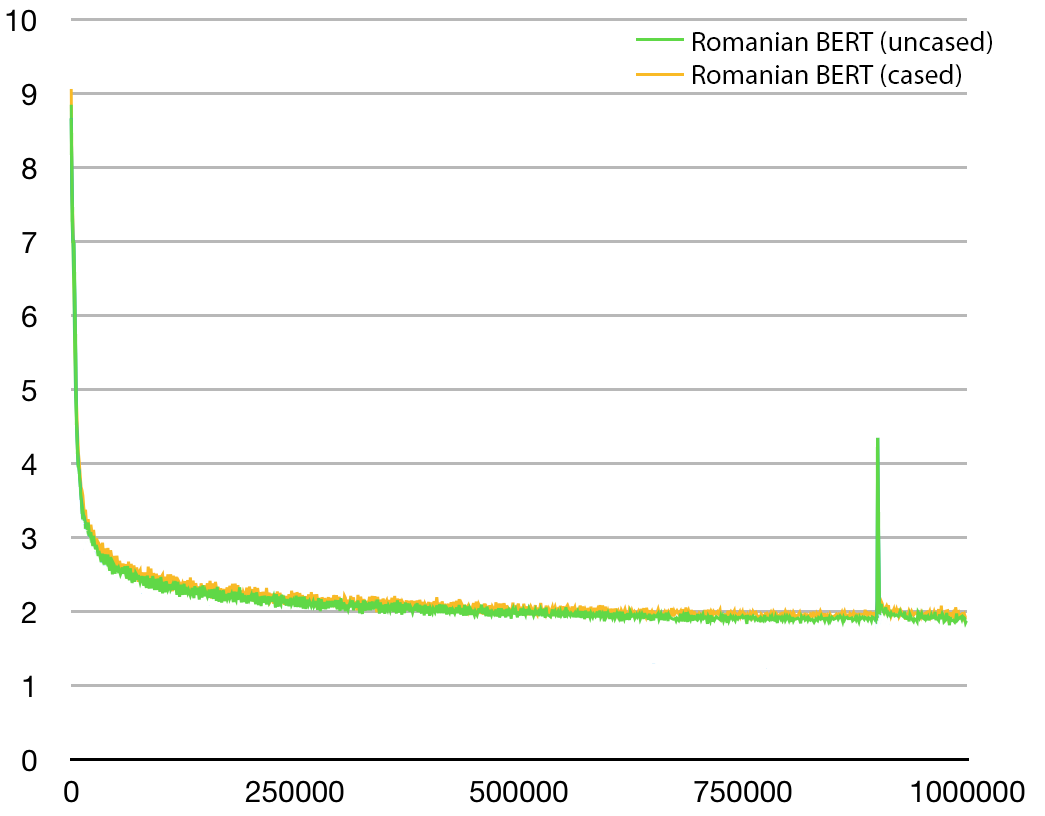}
    \caption{Training loss for the Romanian BERT cased and uncased models.}
    \label{fig:training-graph}
\end{figure}

\begin{table*}
    \centering
    \begin{tabular}{lcccc}
          & \multicolumn{2}{c}{\textbf{Frozen}} & \multicolumn{2}{c}{\textbf{Non-Frozen}} \\
          \multicolumn{1}{l}{\textbf{Model}} & \textbf{UPOS} & \textbf{XPOS} & \textbf{UPOS} & \textbf{XPOS} \\
          \hline
            {\small\verb|M-BERT (uncased)|} & 95.48 & 89.84 & 97.65 & 95.72 \\
            {\small\verb|M-BERT (cased)|} & 94.46 & 89.50 & 97.87 & 96.16 \\
            {\small\verb|Romanian BERT (uncased)|} & \textbf{96.55} & \textbf{95.14} & \textbf{98.18} & \textbf{96.84} \\
            {\small\verb|Romanian BERT (cased)|} & 96.49 & 95.01 & 98.00 & 96.46 \\
          \hline
    \end{tabular}
    \caption{Simple Universal Dependencies evaluation results.}
    \label{tab:rrt_metrics}
\end{table*}

\begin{table*}
    \centering
    \begin{tabular}{lcccc}
          \hline
          \textbf{Model} & \textbf{UPOS} & \textbf{UFeats} & \textbf{Lemmas} & \textbf{LAS} \\
          \hline
            {\small\verb|M-BERT (uncased)|} & 97.72 & 96.54 & 94.67 & 87.65 \\
            {\small\verb|M-BERT (cased)|} & 97.90 & 96.71 & \textbf{95.2} & 88.05 \\
            {\small\verb|Romanian BERT (uncased)|} & \textbf{97.91} & 97.01 & 94.93 & 89.61 \\
            {\small\verb|Romanian BERT (cased)|} & 97.90 & \textbf{97.22} & 94.88 & \textbf{89.69}\\
          \hline
    \end{tabular}
    \caption{Joint Universal Dependencies evaluation results.}
    \label{tab:udify_metrics}
\end{table*}

Vocabulary plays an important part in a the performance of a language model. Broadly speaking, the better sentences are tokenized (roughly, the fewer pieces each word is broken into), the better the model is expected to performs. Comparing {\small\verb|M-BERT|}'s vocabulary with ours (Table~\ref{tab:vocab-stats}), we see that on average, {\small\verb|Romanian BERT|} is able to encode a word in $\sim$1.4 tokens while {\small\verb|M-BERT|} can reach up to 2 tokens/word for the cased vocabulary.
The table also shows that {\small\verb|Romanian BERT|} has an order of magnitude fewer unknown tokens than {\small\verb|M-BERT|} on the same text\footnote{These statistics were computed on a development file extracted from the corpus. This file is composed of 5000 lines that were extracted respecting each individual corpus's size, reflecting a non-skewed distribution. Neither the vocabulary nor the model were trained on this file.}. Table \ref{tab:vocab-example} shows a tokenization example for each evaluated model.

Pretraining was performed using the standard BERT recipe. Each model was trained for 1M steps, with the initial 900K trained on a sequence length of 128 and the rest with the maximum length of 512. Figure~\ref{fig:training-graph} shows the progress for both models. The sudden increase in loss at the 900K steps mark is due to the switch to the 512 sequence length, but both models quickly recover from it. The models were trained using a batch size of 140 per GPU (for 128 sequence length), and then 20 (for 512 sequence length). The optimizer used was Layer-wise Adaptive Moments optimizer for Batch training (LAMB \cite{you2019large}), with warm-up over the first 1\% of steps up to a learning rate of 1e-4, followed by a decay. Eight Nvidia Volta V100 GPUs with 32GB memory were used, and the pretraining process took around 2 weeks per model.

\section{Evaluation} \label{Evaluation}

We evaluate {\small\verb|Romanian BERT|} on three downstream tasks:

\begin{itemize}[noitemsep]
\item \textbf{Simple Universal Dependencies}: one model per task, evaluating labeling performance on UPOS (Universal Part-of-Speech) and the XPOS (eXtended Part-of-Speech).
\item \textbf{Joint Universal Dependencies}: a single model trained jointly on all tasks, evaluating UPOS, UFeats (Universal Features), Lemmas and Dependency Parsing for which we compute the Labeled Attachment Score.
\item \textbf{Named Entity Recognition}: a single model predicting BIO-style labels.
\end{itemize}

For each task we compare {\small\verb|Romanian BERT|} with {\small\verb|M-BERT|}, on both cased and uncased versions. To mitigate the effect of the random seed we run each experiment 5 times and report only the mean score.

All results listed here are reproducible by using the evaluation scripts provided on \href{https://github.com/dumitrescustefan/Romanian-Transformers}{GitHub}. 

\subsection{Simple Universal Dependencies}

\begin{table*}
    \centering
    \begin{tabular}{lcccc}
          \hline
          \textbf{Model} & \textbf{Entity Type} & \textbf{Partial} & \textbf{Strict} & \textbf{Exact} \\
          \hline
            {\small\verb|M-BERT (uncased)|} & 84.75 & 86.06 & 80.81 & 83.91 \\
            {\small\verb|M-BERT (cased)|} & 84.52 & 86.27 & 80.6 & 84.13 \\
            {\small\verb|Romanian BERT (uncased)|} & 85.53 & 87.17 & 82.01 & 85.26 \\
            {\small\verb|Romanian BERT (cased)|} & \textbf{86.21} & \textbf{87.84} & \textbf{82.54} & \textbf{85.88} \\ 
            
          \hline
    \end{tabular}
    \caption{Named entity recognition evaluation results.}
    \label{tab:ronec_metrics}
\end{table*}

For this token labeling task we used the Romanian RRT \cite{barbu2016romanian} dataset from Universal Dependencies (UD), and evaluated the performance of the language models for UPOS and XPOS tagging using the macro-averaged F1, as proposed by \newcite{zeman2018conll}.

\paragraph{Methodology} The model itself is straight-forward: on top of BERT's output layer we directly use a linear layer (with a 0.1 fixed dropout) that projects BERT's outputs into the desired number of classes, depending on the UPOS or XPOS task. Cross-entropy loss is used on the softmaxed linear layer. We perform 2 tests for each UPOS/XPOS task: a frozen test where we train only the added last layer of the model ("freezing" the language model weights), and a full test where all parameters are trainable. The frozen test can give additional insight into a model because its performance now rests more on the pretrained weights rather than the fine-tuned weights, meaning that differences in frozen model performance should be more exaggerated than the fully tuned ones.

\paragraph{Results} The results are summarized in Table \ref{tab:rrt_metrics}. They show that {\small\verb|Romanian BERT|} outperforms {\small\verb|M-BERT|} on all subtasks, with differences in scores ranging from 0.13\% (UPOS non-frozen cased) to 6\% (XPOS frozen cased). Moreover, our assumption that the difference between frozen variants will be higher holds.

Surprisingly, the uncased version of {\small\verb|Romanian BERT|} achieved the highest performance on all experiments, although the cased version should be at something of an advantage on this task thanks to a capital letter indicating proper nouns, etc. We believe these results may be due to better matching between the uncased vocabulary (which necessarily contains more full words) and the RRT corpus.

\subsection{Joint Universal Dependencies}

For this task we evaluated the language models on the same dataset as in the Simple task, namely RRT. We evaluate the models using to the standard CoNLL shared task evaluation tools \cite{zeman2018conll} and report the scores for UPOS (giving us a comparison to the same UPOS task using a simpler model), UFeats (Universal Features of each word), Lemmas, and the Labeled Attachment Score (LAS). 

\paragraph{Methodology} Although we evaluated the models on the same dataset, the methodology was different. We use an external tool to perform evaluation: UDify \cite{kondratyuk201975}, a Transformer-based model that performs joint training and prediction on every UD subtask in a single step. It implements a prediction layer on top of the contextualized embeddings for each task and a layer-wise dot-product attention that calculates a weighted sum for all intermediate representations of a token. 

\paragraph{Results} The results shown in table \ref{tab:udify_metrics} reflect that scores are rather mixed, with the uncased version of {\small\verb|Romanian BERT|} obtaining the highest UPOS score, but with a very small difference from the cased model. For UFeats the {\small\verb|Romanian BERT|} cased is the clear winner with a score of 97.22\%. Somewhat surprisingly, {\small\verb|M-BERT|} cased obtained the highest score on lemma generation. The reasoning for this could be that because {\small\verb|M-BERT|} was trained on multiple languages, including many corpora representing Latin languages (of which Romanian is one), it was able to better model lemmas due to the powerful cross-lingual learning capabilities of the Transformer\footnote{This particular result will lead us to investigate the performance of a multilingual Transformer trained on the Latin languages family}. Finally, the LAS score shows a clear improvement from {\small\verb|M-BERT|} with 87.65\% uncased and 88.05\% cased to {\small\verb|Romanian BERT|} with 89.61\% uncased and 89.69\% cased.

\subsection{Named Entity Recognition}

{\small\verb|Romanian BERT|} was evaluated on RONEC \cite{dumitrescu2019introducing} - a fine-grained NER corpus. The standard BIO tagging was used and models were evaluated according to \cite{segura2013semeval}, reporting the F1 macro-averaged metrics for entity type, partial, strict and exact matches. 

\paragraph{Methodology} The methodology used for this task is identical to the one used for the Simple Universal Dependencies task. However, we do not evaluate the models on their frozen versions.

\paragraph{Results} Table \ref{tab:ronec_metrics} summarizes NER results. Unsurprisingly, we find that {\small\verb|Romanian BERT|} cased has the best performance on this task, improving on {\small\verb|M-BERT|} scores with 0.78\% to 1.96\% on all metrics. The uncased version of {\small\verb|Romanian BERT|} placed second at the evaluation, outperforming both cased and uncased versions of {\small\verb|M-BERT|}.

\section{Conclusions}
We have introduced the first BERT model trained solely on a 15GB Romanian corpus, obtained by a thorough clean of OSCAR, OPUS and Wikipedia sub-corpora. We have shown that {\small\verb|Romanian BERT|} outperforms the only available model that works on Romanian, the multilingual {\small\verb|M-BERT|}.

As the current corpus is better cleaned, more text added, tweaks to improve vocabulary coverage are performed, new versions of BERT as well as future models will be released to the open source domain\footnote{\url{ https://github.com/dumitrescustefan/Romanian-Transformers}}.

\bibliography{emnlp2020}

\begin{thebibliography}{22}
\expandafter\ifx\csname natexlab\endcsname\relax\def\natexlab#1{#1}\fi

\bibitem[{Barbu~Mititelu et~al.(2016)Barbu~Mititelu, Ion, Simionescu, Irimia,
  and Perez}]{barbu2016romanian}
Verginica Barbu~Mititelu, Radu Ion, Radu Simionescu, Elena Irimia, and
  Cenel-Augusto Perez. 2016.
\newblock The romanian treebank annotated according to universal dependencies.
\newblock In \emph{Proceedings of the tenth international conference on natural
  language processing (hrtal2016)}.

\bibitem[{Clark et~al.(2019)Clark, Khandelwal, Levy, and
  Manning}]{clark2019does}
Kevin Clark, Urvashi Khandelwal, Omer Levy, and Christopher~D. Manning. 2019.
\newblock \href {http://arxiv.org/abs/1906.04341} {What does bert look at? an
  analysis of bert's attention}.

\bibitem[{Clark et~al.(2020)Clark, Luong, Le, and Manning}]{clark2020electra}
Kevin Clark, Minh-Thang Luong, Quoc~V. Le, and Christopher~D. Manning. 2020.
\newblock \href {http://arxiv.org/abs/2003.10555} {Electra: Pre-training text
  encoders as discriminators rather than generators}.

\bibitem[{Devlin et~al.(2018)Devlin, Chang, Lee, and
  Toutanova}]{devlin2018bert}
Jacob Devlin, Ming-Wei Chang, Kenton Lee, and Kristina Toutanova. 2018.
\newblock Bert: Pre-training of deep bidirectional transformers for language
  understanding.
\newblock \emph{arXiv preprint arXiv:1810.04805}.

\bibitem[{Dumitrescu and Avram(2019)}]{dumitrescu2019introducing}
Stefan~Daniel Dumitrescu and Andrei-Marius Avram. 2019.
\newblock Introducing ronec--the romanian named entity corpus.
\newblock \emph{arXiv preprint arXiv:1909.01247}.

\bibitem[{Kondratyuk(2019)}]{kondratyuk201975}
Daniel Kondratyuk. 2019.
\newblock 75 languages, 1 model: Parsing universal dependencies universally.
\newblock \emph{arXiv preprint arXiv:1904.02099}.

\bibitem[{Lample and Conneau(2019)}]{lample2019crosslingual}
Guillaume Lample and Alexis Conneau. 2019.
\newblock \href {http://arxiv.org/abs/1901.07291} {Cross-lingual language model
  pretraining}.

\bibitem[{Lan et~al.(2019)Lan, Chen, Goodman, Gimpel, Sharma, and
  Soricut}]{lan2019albert}
Zhenzhong Lan, Mingda Chen, Sebastian Goodman, Kevin Gimpel, Piyush Sharma, and
  Radu Soricut. 2019.
\newblock \href {http://arxiv.org/abs/1909.11942} {Albert: A lite bert for
  self-supervised learning of language representations}.

\bibitem[{Le et~al.(2019)Le, Vial, Frej, Segonne, Coavoux, Lecouteux, Allauzen,
  Crabbé, Besacier, and Schwab}]{le2019flaubert}
Hang Le, Loïc Vial, Jibril Frej, Vincent Segonne, Maximin Coavoux, Benjamin
  Lecouteux, Alexandre Allauzen, Benoît Crabbé, Laurent Besacier, and Didier
  Schwab. 2019.
\newblock \href {http://arxiv.org/abs/1912.05372} {Flaubert: Unsupervised
  language model pre-training for french}.

\bibitem[{Michel et~al.(2019)Michel, Levy, and Neubig}]{michel2019sixteen}
Paul Michel, Omer Levy, and Graham Neubig. 2019.
\newblock \href {http://arxiv.org/abs/1905.10650} {Are sixteen heads really
  better than one?}

\bibitem[{Ortiz~Su{\'a}rez et~al.(2019)Ortiz~Su{\'a}rez, Sagot, and
  Romary}]{ortizsuarez:hal-02148693}
Pedro~Javier Ortiz~Su{\'a}rez, Beno{\^i}t Sagot, and Laurent Romary. 2019.
\newblock \href {https://hal.inria.fr/hal-02148693} {{Asynchronous Pipeline for
  Processing Huge Corpora on Medium to Low Resource Infrastructures}}.
\newblock In \emph{{7th Workshop on the Challenges in the Management of Large
  Corpora (CMLC-7)}}, Cardiff, United Kingdom.

\bibitem[{Rogers et~al.(2020)Rogers, Kovaleva, and
  Rumshisky}]{rogers2020primer}
Anna Rogers, Olga Kovaleva, and Anna Rumshisky. 2020.
\newblock \href {http://arxiv.org/abs/2002.12327} {A primer in bertology: What
  we know about how bert works}.

\bibitem[{Sanh et~al.(2019)Sanh, Debut, Chaumond, and
  Wolf}]{sanh2019distilbert}
Victor Sanh, Lysandre Debut, Julien Chaumond, and Thomas Wolf. 2019.
\newblock \href {http://arxiv.org/abs/1910.01108} {Distilbert, a distilled
  version of bert: smaller, faster, cheaper and lighter}.

\bibitem[{Segura~Bedmar et~al.(2013)Segura~Bedmar, Mart{\'\i}nez, and
  Herrero~Zazo}]{segura2013semeval}
Isabel Segura~Bedmar, Paloma Mart{\'\i}nez, and Mar{\'\i}a Herrero~Zazo. 2013.
\newblock Semeval-2013 task 9: Extraction of drug-drug interactions from
  biomedical texts (ddiextraction 2013).
\newblock Association for Computational Linguistics.

\bibitem[{Tenney et~al.(2019)Tenney, Das, and Pavlick}]{tenney2019bert}
Ian Tenney, Dipanjan Das, and Ellie Pavlick. 2019.
\newblock \href {http://arxiv.org/abs/1905.05950} {Bert rediscovers the
  classical nlp pipeline}.

\bibitem[{Tiedemann(2012)}]{Tiedemann2012ParallelDT}
J{\"o}rg Tiedemann. 2012.
\newblock Parallel data, tools and interfaces in opus.
\newblock In \emph{LREC}.

\bibitem[{Vaswani et~al.(2017)Vaswani, Shazeer, Parmar, Uszkoreit, Jones,
  Gomez, Kaiser, and Polosukhin}]{vaswani2017attention}
Ashish Vaswani, Noam Shazeer, Niki Parmar, Jakob Uszkoreit, Llion Jones,
  Aidan~N Gomez, {\L}ukasz Kaiser, and Illia Polosukhin. 2017.
\newblock Attention is all you need.
\newblock In \emph{Advances in neural information processing systems}, pages
  5998--6008.

\bibitem[{Virtanen et~al.(2019)Virtanen, Kanerva, Ilo, Luoma, Luotolahti,
  Salakoski, Ginter, and Pyysalo}]{virtanen2019multilingual}
Antti Virtanen, Jenna Kanerva, Rami Ilo, Jouni Luoma, Juhani Luotolahti, Tapio
  Salakoski, Filip Ginter, and Sampo Pyysalo. 2019.
\newblock \href {http://arxiv.org/abs/1912.07076} {Multilingual is not enough:
  Bert for finnish}.

\bibitem[{de~Vries et~al.(2019)de~Vries, van Cranenburgh, Bisazza, Caselli, van
  Noord, and Nissim}]{vries2019bertje}
Wietse de~Vries, Andreas van Cranenburgh, Arianna Bisazza, Tommaso Caselli,
  Gertjan van Noord, and Malvina Nissim. 2019.
\newblock \href {http://arxiv.org/abs/1912.09582} {Bertje: A dutch bert model}.

\bibitem[{Yang et~al.(2019)Yang, Dai, Yang, Carbonell, Salakhutdinov, and
  Le}]{yang2019xlnet}
Zhilin Yang, Zihang Dai, Yiming Yang, Jaime Carbonell, Ruslan Salakhutdinov,
  and Quoc~V. Le. 2019.
\newblock \href {http://arxiv.org/abs/1906.08237} {Xlnet: Generalized
  autoregressive pretraining for language understanding}.

\bibitem[{You et~al.(2019)You, Li, Reddi, Hseu, Kumar, Bhojanapalli, Song,
  Demmel, Keutzer, and Hsieh}]{you2019large}
Yang You, Jing Li, Sashank Reddi, Jonathan Hseu, Sanjiv Kumar, Srinadh
  Bhojanapalli, Xiaodan Song, James Demmel, Kurt Keutzer, and Cho-Jui Hsieh.
  2019.
\newblock Large batch optimization for deep learning: Training bert in 76
  minutes.
\newblock In \emph{International Conference on Learning Representations}.

\bibitem[{Zeman et~al.(2018)Zeman, Hajic, Popel, Potthast, Straka, Ginter,
  Nivre, and Petrov}]{zeman2018conll}
Daniel Zeman, Jan Hajic, Martin Popel, Martin Potthast, Milan Straka, Filip
  Ginter, Joakim Nivre, and Slav Petrov. 2018.
\newblock Conll 2018 shared task: Multilingual parsing from raw text to
  universal dependencies.
\newblock In \emph{Proceedings of the CoNLL 2018 Shared Task: Multilingual
  parsing from raw text to universal dependencies}, pages 1--21.

\end{thebibliography}
\bibliographystyle{acl_natbib}

\end{document}